\documentclass[sigconf]{acmart}
\pdfoutput=1
\renewcommand\footnotetextcopyrightpermission[1]{} 
\usepackage{booktabs}
\usepackage{todonotes}
\usepackage{xcolor}
\usepackage{algorithm2e}
\usepackage{graphicx}
\usepackage{subcaption}
\usepackage{arydshln}
\usepackage{dsfont}
\usepackage{datetime}
\usepackage{multirow}
\usepackage{longtable}

\newdateformat{monthyeardate}{%
\monthname[\THEMONTH], \THEYEAR}

\AtBeginDocument{%
  \providecommand\BibTeX{{%
    \normalfont B\kern-0.5em{\scshape i\kern-0.25em b}\kern-0.8em\TeX}}}

\acmYear{2019}
\setcopyright{none}
\acmConference[ACM FAT* 2020]{ACM FAT* Conference 2020: ACM Conference on Fairness, Accountability, andTransparency}{January 27--30, 2020}{Barcelona, Spain}

\begin{document}


\title[Recommendation or Discrimination?] {Recommendation or Discrimination?: Quantifying Distribution Parity in Information Retrieval Systems}



\author{Rinat Khaziev}
\email{rkhaziev@truefit.com}
\affiliation{%
 \institution{True Fit Corporation}
 \city{Boston}
 \state{MA}
}

\author{Bryce Casavant}
\email{bcasavant@truefit.com}
\affiliation{%
	\institution{True Fit Corporation}
	\city{Boston}
	\state{MA}
}

\author{Pearce Washabaugh}
\email{pwashabaugh@truefit.com}
\affiliation{%
	\institution{True Fit Corporation}
	\city{Boston}
	\state{MA}
}

\author{Amy A. Winecoff}
\email{awinecoff@truefit.com}
\affiliation{%
	\institution{True Fit Corporation}
	\city{Boston}
	\state{MA}
}
\author{Matthew Graham}
\email{mgraham@truefit.com}
\affiliation{%
	\institution{True Fit Corporation}
	\city{Boston}
	\state{MA}
}

\renewcommand{\shortauthors}{XXX, et al.}

\begin{abstract}

Information retrieval (IR) systems often leverage query data to suggest relevant items to users. This introduces the possibility of unfairness if the query (i.e., input) and the resulting recommendations unintentionally correlate with latent factors that are protected variables (e.g., race, gender, and age). For instance, a visual search system for fashion recommendations may pick up on features of the human models rather than fashion garments when generating recommendations. In this work, we introduce a statistical test for "distribution parity" in the top-K IR results, which assesses whether a given set of recommendations is fair with respect to a specific protected variable. We evaluate our test using both simulated and empirical results. First, using artificially biased recommendations, we demonstrate the trade-off between statistically detectable bias and the size of the search catalog. Second, we apply our test to a visual search system for fashion garments, specifically testing for recommendation bias based on the skin tone of fashion models. Our distribution parity test can help ensure that IR systems' results are fair and produce a good experience for all users. 

\end{abstract}

%

\begin{teaserfigure}
    \centering
  \includegraphics[width=\textwidth]{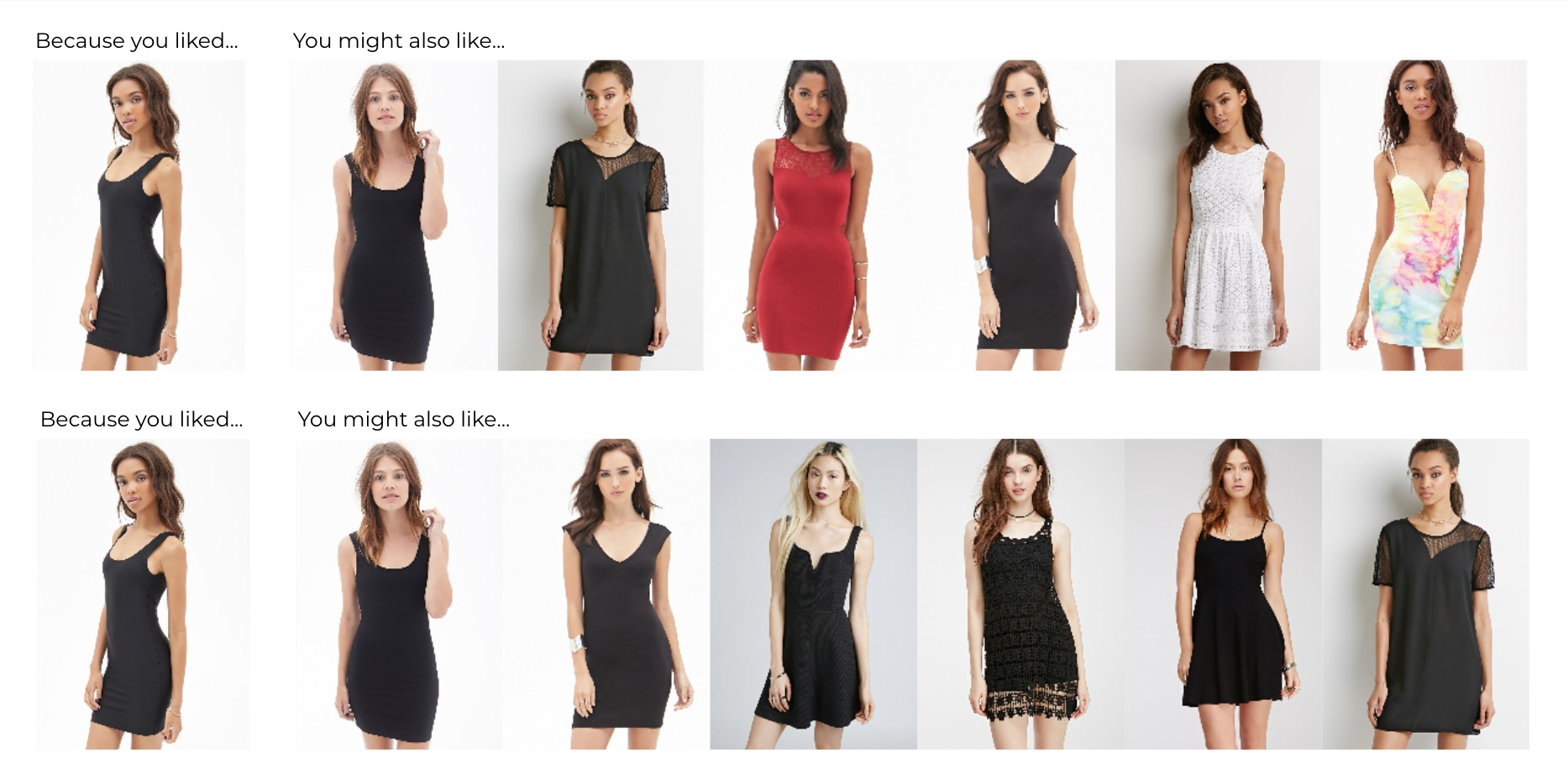}
  \caption{(Above) Biased recommendations; the recommendations pick up on the query fashion model rather than the fashion garment. (Below) Unbiased recommendations; the recommendations pick up on the query fashion garment not the fashion model.}
  \label{fig:teaser}
\end{teaserfigure}

\maketitle

\section{Introduction}

Information retrieval (IR) systems, such as search engines and recommender systems (RS), are some of the most widely used machine learning systems today and are used to suggest a list of items or recommendations that are most relevant to users. Given the widespread use of IR systems and RS, ensuring that all users receive the same, high quality recommendation experience is critically important. Unlike other applications of machine learning, IR systems make recommendations based on a query or input. If the algorithms used to produce recommendations have underlying biases, these biases will likely propagate to the results. More specifically, if algorithms used in IR systems have encoded information associated with latent protected variables such as gender, race, or age, the presence of a protected characteristic (e.g., being a woman) in the query can lead to recommendation results also reflecting this protected characteristic. In most domains of application, protected variables are not related to the features that matter most for recommendations. For example, a user's race is not relevant to whether or not two handbags are similar. Yet if race were partially encoded in the query handbag, for example through characteristics of a human model in an image as well as in the IR algorithm, recommendations may also reflect this bias. Being able to quantify the extent to which recommendations reflect irrelevant, protected variables is a necessary first step in ensuring that all users receive fair and useful recommendations. 

The risk of generating unfair recommendations is especially acute for applications like visual search that use features extracted from images to generate recommendations. In many modern visual search systems, deep neural networks learn the visual features used to identify relevant recommendations. However, the standard datasets used to train deep neural networks have implicit gender and racial biases \cite{handa_2019}. These biases become encoded in the recommendation models themselves, leading to discriminatory and biased results \cite{stock2018convnets}. In fashion visual search applications, human fashion models often appear in the query image and result images along with the fashion garments being recommended. Therefore, using a biased computer vision model can lead to recommendations that have more in common with the people modeling fashion garments, such as the dress images in the top row of Figure \ref{fig:teaser}, than the fashion garments themselves, such as the dress images in the bottom row of Figure \ref{fig:teaser}. 

In recent years, fairness in machine learning has received increased attention both publicly and in the scientific community, leading to some convergence in operational definitions of fairness. Definitions of fairness typically fall into two categories: individual fairness and group fairness \cite{Dwork2012}, \cite{yang2017measuring}, \cite{Zehlike2017}, \cite{Karako2018}, and \cite{Gajane2017}. Individual fairness is achieved when everyone is treated consistently regardless of their association with a protected group or protected variable. Group fairness, also known as statistical parity or demographic parity, requires that a group receiving a positive or negative outcome is treated equal to all other groups. That is, group fairness is achieved when outcomes are equalized across all groups. These definitions have provided a conceptual framework for researching fairness in many machine learning systems; however, they are not directly applicable in all contexts. For example, in IR, the output of a model is not a single, categorical determination, but rather a list of recommendations. For IR to be fair, lists of recommendations must be independent of protected variables. 

In this paper, we introduce a new definition of fairness, \textquotedblleft distribution parity,\textquotedblright{} to assess the fairness of IR systems. An IR system exhibits distribution parity when the distribution of values of a protected variable in the top-K recommendations match the distribution of values in the dataset regardless of the value of the protected variable in the query. In contrast, an IR system lacks distribution parity when the value of the protected variable in the input significantly biases the distribution of values of the protected variable in the top-K recommendations relative to the dataset. For example, if an IR system returns more images of women relative to the dataset when the query image contains a woman, the system would fail to satisfy distribution parity.  

Using the concept of distribution parity, we develop an approach to determine whether a IR system's recommendations are biased and evaluate our approach in the context of fashion recommendations. We first describe our statistical test for distribution parity. We then use Monte Carlo simulations to investigate the relationship between the sample size, bias, and statistical power of our distribution parity test. Lastly, we apply our test to a fashion visual search system to elucidate how the test could function in a real-world context. Specifically, we use a neural network based visual search system that uses image embeddings to retrieve similar clothing items. For images, we use the DeepFashion In-Shop Clothes Retrieval \cite{liuLQWTcvpr16DeepFashion} dataset. Using image segmentation, we extract the skin tone of the fashion models depicted in the DeepFashion images and apply our distribution parity test to determine if, given an image with a fashion model with a particular skin tone, the resulting similar images are significantly more or less likely to contain fashion models with the same skin tone. 

\section{Related Work on Fairness in Machine Learning}
Much of the machine learning bias research focuses on discrimination in classification problems, Verma and Rubin \cite{Verma2018} have a good review of what has been done; however, our statistical test focuses on bias in recommendation problems. Only a handful of papers look at bias in recommendation problems \cite{yang2017measuring}, \cite{Karako2018}, and \cite{Zehlike2017}. Of the recommendation focused papers, Yang and Stonyanovich \cite{yang2017measuring} explicitly measures any form of bias in RS. They define fairness as statistical parity where the proportion of the protected group in the top-K recommendations is the same as the non-protected group and use modified ranking metrics to detect fairness in ranked lists. To test for statistical parity with multiple protected groups, they use a modified KL-divergence, normalized discounted KL-divergence (rKL). A major drawback of rKL is that it does not give a clear answer for when a ranking algorithm returns biased recommendations. Other papers that have investigated biases in RS have focused on inclusivity in results, but do not give explicit measures. Karako and Manggala \cite{Karako2018} define fairness as uniformity of the protected variables in the top-K recommendations. Such an approach yields a recommender system that is perceptually equal, however, it can over-recommend items associated with the minority labels regardless of their relevance to the user. Zehlike et. al. \cite{Zehlike2017} utilizes statistical parity with the criterion that the proportion of a protected group is above a minimum threshold, $p$, set by the practitioner. Previous methods for defining fairness in recommendations have primarily focused only on the recommendation results, yet results in many IR systems are contingent upon some input, such as a query image in a visual search system. In the current work, our test for distribution parity addresses this problem by defining fairness in results with respect to an input.

\section{Approach}
\label{sec:approach}
Our approach consists of three parts: 1) a statistical test for distribution parity; 2) an evaluation of the statistical power of the test using simulated data; and 3) an application of the test to a real-world dataset using visual search for fashion recommendations.  

\subsection{Statistical Test for Distribution Parity}
\subsubsection{Definitions}

In this work, we consider recommendations unbiased if the query-conditional distribution exhibits distribution parity. In a dataset, there is a protected variable $Z$ (e.g., skin tone) with a set of possible values where $z \in Z$ is a given value of the protected variable (e.g., ST3 skin tone). An individual observation in the dataset can only be associated with one value of the protected variable. The distribution of the values of the protected variable, conditional on the query input's value, $z^q$, is $P_Z(z|z^q)$. Generally, to achieve distribution parity, the distribution of values of the protected variable in the recommendations should on average match the distribution $P_Z(z)$ of the protected variable values in the dataset. 

We can concretely enforce distribution parity in two ways, either for each of the top-K ranks $r$
\begin{equation}
P_{Z^k}\left(z|z^q, r = k\right) = P_Z\left(z\right), \forall k < K
\label{eq:strong-fairness}
\end{equation}
or for a set of top-K ranks
\begin{equation}
P_{Z^{<K}}\left(z|z^q, r < K\right) = P_Z\left(z\right),
\label{eq:weak-fairness}
\end{equation}
where $Z^k$ and $Z^{<k}$ denote the random variables associated with the evaluation of $Z$ over positions $k$ and top-$K$ of recommendations respectively.

We refer to equations \eqref{eq:strong-fairness} and \eqref{eq:weak-fairness} as strong and weak fairness conditions. Detecting strong fairness is preferred when end users are exposed to a large set of ranked outputs. Enforcing the weak fairness condition is usually sufficient especially with fashion IR systems since they typically do not display more than a handful of items to each user and deal with relatively small search catalogs. 

If recommendations lack distribution parity, the distribution of values of the protected variable in the set of recommendations will be significantly different than the distribution of the values of the protected variable in the dataset. Such a definition of fairness is particularly applicable when the protected variable's distribution of values is highly imbalanced (e.g., skin tone of human models in fashion catalogs), ensuring that underrepresented protected variable values are not diluted during evaluation of recommendations.

\subsubsection{Testing Fairness using a Categorical Protected Variable}\label{sec:testing-fairness}
\label{sec:test-definitions}

\begin{table}[htbp]
	\small
	\centering
	\caption{Example of a omnibus contingency tables for detecting bias in the recommendations for a catalog with three values (A, B, and C) of a protected variable, $K=6$}
	\begin{tabular}{cccc}
		\toprule
		\multirow{2}{*}{Query Items} & \multicolumn{3}{c}{Recommendations} \\ \cline{2-4}
		& $z=A$ & $z=B$ & $z=C$	\\ \midrule
		$z^q=A$ & 300	& 50 & 250 \\
		$z^q=B$  & 40 & 600 & 260 \\
		$z^q=C$ 	&  80 & 150 & 1800 \\
		catalog	& 100 & 150 & 350 \\
		\bottomrule
	\end{tabular}
	\label{tab:contidency-table1}
\end{table}

We verify the statistical validity of condition \eqref{eq:weak-fairness} by performing a test of independence on a contingency table (see Table \ref{tab:contidency-table1}) that is generated by aggregating the top-K recommendations for each value of the protected variable. Each row of the table encodes the protected variable's distribution of values (e.g., skin tones), $z$, in the top-K recommendations for a given value $z^q$. To test recommendations for distribution parity, we include the protected variable's distribution of values in the search catalog as a row of the contingency table. We test for the independence (i.e., distribution parity) of the protected variable using a $\chi^2$-test, which is performed on the full contingency table for a given $\alpha$ level. We refer to the test using the full contingency table as the \textquotedblleft omnibus test\textquotedblright. The null hypothesis $H_0^O$ of the omnibus test is formulated as follows
\begin{equation}
H_0^O:\ P_{Z^{<K}}(z | z^q, r < K) = P(z^q),\ \forall z,\ \forall z^q,
\label{eq:global-null-hypothesis}
\end{equation}

If the omnibus test detects a statistically significant effect, it may be of interest to determine whether there is a significant effect for specific values of the protected variable (e.g., for the skin tone ST1 or ST3). We refer to these follow up tests, focused on a single value of the protected variable, as \textquotedblleft contrast tests\textquotedblright. The contrast test is performed by setting up a contingency table for a single value of $z^q$ along with the protected variable's distribution of values in the search catalog (see Table \ref{tab:contidency-table2}). The null hypothesis of a contrast test $H_0^C(z^q)$ is defined for every value of the protected variable $z^q$ as
\begin{equation}
H_0^C(z^q):\ P_{Z^{<K}}(z^q | z^q,  r < k) = P(z^q), \forall z^q.
\label{eq:local-null-hypothesis}
\end{equation}
In the contrast test, we use a $2 \times 2$ aggregation of the full contingency table for each value of the protected variable. Operating on a subset of the full table for the value of interest, the contrast test \eqref{eq:local-null-hypothesis} has less power than test omnibus test  \eqref{eq:global-null-hypothesis}, and as a result, there is a risk of not detecting biases on small datasets.

\begin{table}[htbp]
	\small
	\centering
	\caption{Example of a contrast contingency table for detecting bias in recommendations for the $z^q=\text{A}$, $K=6$}
	\begin{tabular}{ccc}
		\toprule
		\multirow{2}{*}{Query Items} & \multicolumn{2}{c}{Recommendations} \\ \cline{2-3}
		            & $z=A$ & $z\neq A$ \\ \midrule
	    $z^q=$A     & 300   & 300 \\
		catalog     & 100   & 600 \\
		\bottomrule
	\end{tabular}
	\label{tab:contidency-table2}
\end{table}

We measure the bias in recommendations for each skin tone $z^q$ using the risk ratio
\begin{equation}
RR (z^q) = \frac{P_{Z^{<K}}\left(z^q|z^q, r < k\right)}{P_Z\left(z^q\right)},
\label{eq:risk-ratio}
\end{equation}
Using the risk ratio is a preferable due to its ease of interpretation. Values of the risk ratio, $RR (z^q)$, greater than one indicate the algorithm is over-representing a protected variable in the search results, and values less than one indicate that a protected variable is under-represented in the search results. In addition to the point estimate for the risk ratio, we can calculate a confidence interval using standard practices. 

It is important to note that a bias can manifest as either an under representation or an over representation of a given value of a protected variable. To quantify a bias in either direction on a common scale, we use a normalized risk ratio
\begin{equation}
nRR (z^q) =
    \begin{cases}
        RR (z^q) & \text{if } RR(z^q) \leq 1 \\
        1/RR (z^q) & \text{otherwise},
\end{cases}       
\label{eq:normalize-risk-ratio}
\end{equation}
which takes values between 0 and 1. The normalized risk ratio \eqref{eq:normalize-risk-ratio} can be linked to the "80\% rule," which is used in the American legal system to define a threshold for discriminatory policies. The 80\% rule specifies that an algorithm can be considered discriminatory if the rate at which individuals belonging to a protected group (e.g., having a disability) are assigned to a positive outcome (e.g., being hired) is less than 80\% of the rate at which individuals not belonging to that group (e.g., not having a disability) are assigned to the positive outcome \cite{zafar2015fairness}. Under the 80\% rule, an algorithm is considered to be fair if the normalized risk ratio $nRR$ belongs to the range from 0.8 to 1 (or risk ratio $RR$ belongs to the range from 0.8 and 1.25).

\subsection{Evaluation of Statistical Power with Simulated Data}
Whether a deviation from distribution parity is statistically detectable depends on both the risk ratio (i.e., bias size) and the sample size (catalog size). To provide an understanding of how the the risk ratio and catalog size impact the likelihood of our omnibus and contrast tests detecting a bias, we perform an analysis using Monte Carlo simulated data.

\label{sec:mc-simulation}
In order to understand the statistical properties of our tests under different conditions, it is necessary to generate synthetic datasets with a known level of bias. Specifically, we generate synthetic datasets with a given distribution of values of the protected variable $P_Z$, artificially manipulating the risk ratio $RR$ and catalog size $N$. When generating synthetic recommendations, we randomly sample $n$ query items from a given distribution of the protected variable $P_Z$, and then generate the top-K recommendations for each item from a biased distribution of the protected variable controlled by the risk ratio $RR$. The aforementioned quantities are used as input variables to our Monte Carlo sampling algorithm, Algorithm \ref{alg:monte-carlo-sampling}. 

Algorithm \ref{alg:monte-carlo-sampling} includes four steps. First, we sample the protected variable values for the query items $Z^q$ from the protected variable's distribution of values $P_Z$. Second, for each query item $q$, we skew the protected variable's distribution of values $P_Z^*$ by multiplying $P_Z(z^q)$ by the risk ratio $RR$, ensuring that $P_Z^*(z^q)$ is no greater than 1. The rest of the probabilities $P_Z^*$ are scaled down uniformly to make sure that  $\sum_{z \in Z}P_Z^*(z)$ = 1 and $\forall P_Z^*(z) \in \left[0,1\right]$. Finally, the $k$ recommendations are sampled from $P_Z^*$ and assigned to the query item $q$.

\begin{algorithm}
\SetAlgoLined
\SetKwInOut{Input}{input}\SetKwInOut{Output}{output}
\Input{$n$, $RR$, $P_Z$, $k$}
\Output{$Q$ query items, $R$ Set of Recommendations}\
initialize\;
$Z^q \leftarrow n \text{ samples from } P_Z$\;
$recommendations \leftarrow \emptyset$\;
\For{$z^q \in Z$}{
    $P_Z^*(z^q) \leftarrow \min \left(1, \ RR \times P_Z\left(z^q\right) \right)$\;
    \For{$z \in Z \neq z^q $}{
        $P_Z^*(z) \leftarrow \left(\frac{1-P_Z^*(z^q)}{1-P_Z(z)}\right) \times P_Z(z)$\;
    }
    $recommendations[z^q] \leftarrow k \text{ samples from } P_Z^*$ \;}
\Return{$recommendations$};
\caption{Monte Carlo sampling algorithm for generating biased recommendations as a function of catalog size $n$, risk ratio $RR$, protected variable's distribution of values $P_Z$, and number of recommendations $k$.}
\label{alg:monte-carlo-sampling}
\end{algorithm}

For each set of simulation parameters, the generated values of the protected variables for the queries and recommendations are used to build the omnibus and contrast contingency tables to evaluate the hypotheses $H_0^O$ and $H_0^C(z^q)$ for $\forall z^q$. Knowing that the alternative hypothesis is true in any of those cases, except for $RR=1$, we empirically evaluate our test's power, a probability of correctly rejecting $H_0$ on $m$ Monte Carlo trials as a fraction of the tests that were rejected out of $M$ trials.

Note that the statistical properties of the hypothesis tests that are generated using this approach are valid only for the protected variable's distribution of values, $P_Z$, specific to a particular search catalog. Thus, a practitioner would have to recompute the statistical properties of the tests every time the protected variable's distribution of values significantly changes in the search catalog. Nevertheless, our simulation analysis can provide general guidance to practitioners for the conditions under which the distribution parity test is appropriate.

\subsection{Application to Visual Search for Fashion Recommendations}

To provide an example of how our approach would function within a real-world information retrieval system, we apply our test for distribution parity within the context of a fashion visual search system. In this analysis, we focus on evaluating distribution parity in the skin tone of fashion models depicted in recommendations. To apply our test in a fashion visual search setting involves several steps. First, we generate recommendations for each image in the dataset using image embeddings learned from a convolutional neural network (CNN) (see Section \ref{sec:visual-search}). Second, we extract the skin tone of the fashion models in each image, which occurs in two stages. In the first stage, we use supervised image segmentation to localize the models' skin within each image (Section \ref{sec:image-segmentation}). In the second stage, we take each skin cutout and extract the skin tone for each image using a tailored color mapping (see Section \ref{sec:skin-tone-extraction}). Lastly, we apply our test for distributional parity to determine whether there is a statistically significant bias in the recommendations using the omnibus test and perform follow up contrasts to elucidate the specific nature of any significant bias detected by the omnibus test. 

Importantly, distributional parity in skin tone is not equivalent to parity in race. We focus specifically on skin tone for several reasons. In many datasets used for classification problems, individuals\textquotesingle{} self-reported race is known. For fashion images, we do not know how fashion models would describe their own race. Without these self-identifications, we cannot determine the race of fashion models because race is a social construct without a fixed meaning \cite{omi2014}. Although race, for the purposes of scientific inquiry, cannot be precisely defined based on objectively quantifiable characteristics \cite{lee2009race}, psychological research has shown that humans use perceptual features such as a person\textquotesingle s skin tone to make racial categorizations \cite{Dunham2015, Stepanova2009}. In addition to being used as a perceptual proxy for race, skin tone can also account for biases that extend beyond race in contexts such as electoral decision making \cite{weaver2012electoral}, implicit attitudes \cite{Nosek2007}, and the marriage marketplace \cite{jha2009looking}. Therefore, although skin tone is not equivalent to race, skin tone is still an important avenue of inquiry for understanding both intra- and inter-racial bias.

\subsubsection{Visual Search}
\label{sec:visual-search}

Visual search systems commonly rely on CNNs to automatically learn and generate features in a high-dimensional space \cite{Krizhevsky2012}. The generated features, referred to as image embeddings, are fixed-length vector image representations from the CNNs hidden layers. However, the image embeddings are not human interpretable. Therefore, protected variables are not easily detectable if CNNs implicitly learn them, hence the need for our test.

Following \cite{Jing2015, Shankar2017, Yang2017}, we build a visual search system that retrieves similar items using image embeddings. The most similar images are determined using a k-nearest neighbor search with the Minkowski distance metric. We utilize a ResNet model \cite{He2016} pre-trained on ImageNet\cite{imagenet_cvpr09} to generate image embeddings for fashion images.

\subsubsection{Image Segmentation for Skin Detection}
\label{sec:image-segmentation}

\begin{figure*}[htbp]
    \centering
    \includegraphics[width=\linewidth]{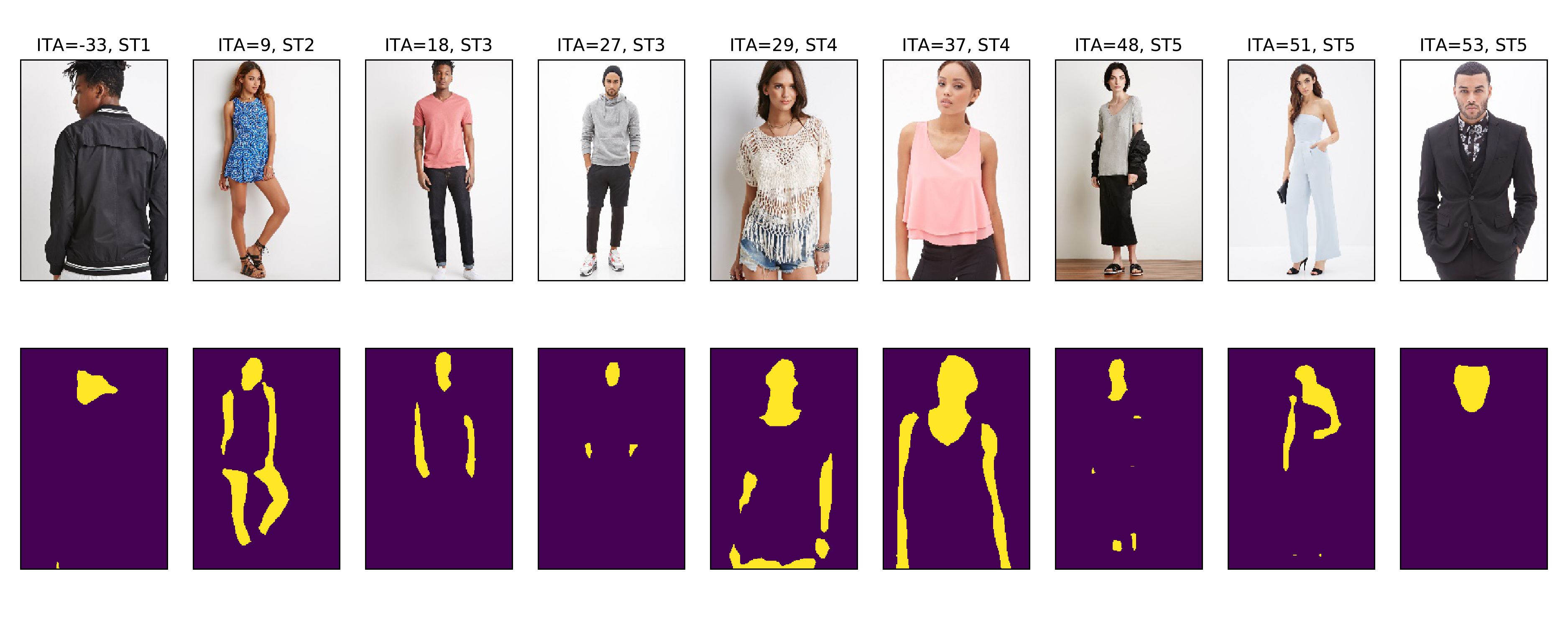}
    \caption{Examples of the skin segmentation and ITA predictions on DeepFashion dataset \cite{liuLQWTcvpr16DeepFashion}. Top row, the original images with ITA values and ITA category listed above. Bottom row, pixel-wise skin detection on images where yellow indicates the areas of the image labeled as skin.}
   \label{fig:segmentation-ita-example}
\end{figure*}

To determine the skin tone of human models in fashion images, we must first extract models\textquotesingle{} skin from the rest of the image. To this end, we train a binary supervised image segmentation model that can distinguish skin from the rest of the image. We use a CNN to perform supervised image segmentation, a task of pixel-wise image classification. Supervised image segmentation is an active field of study \cite{Chen2018,Chen2018cv}, and has many fashion applications \cite{Yang2014, Liu2014, Liu2016, Tangseng2017}. In this work, we use the DeepLab V3+ architecture with the Xception feature extractor and output stride 16 \cite{Chen2018cv}.

 We trained our image segmentation model on a proprietary dataset that includes images that are representative of online fashion catalogs. Although there are publicly available datasets for training image segmentation models on fashion images, they do not provide labels that are sufficiently accurate to train a model for skin tone extraction. The largest publicly-available fashion segmentation dataset, ModaNet \cite{Zheng2018}, does not have skin labels at all. The Fashionista\cite{Yang2014} and Clothing Co-parsing\cite{Tangseng2017} datasets provide skin labels; however, they were generated using super-pixel labeling techniques and as a result have noisy label boundaries with a large number of false positive pixels.

\subsubsection{Skin Tone Classification}
\label{sec:skin-tone-extraction}

Given a trained image segmentation model, we are able to extract skin pixels; however, we need a framework for classifying the skin tone of skin pixels to perform our test. Extracting the skin tone using raw RGB values (16M colors) or color histograms results in feature spaces that do not map well to human perceptions of skin tone. In dermatology research, skin tones are often characterized using Fitzpatrick \cite{Fitzpatrick1975, Fitzpatrick1988} skin type and, more recently, Individual Typology Angle (ITA) \cite{chardon1991skin}.

The Fitzpatrick skin type is determined based on a self-reported questionnaire, and consequently, cannot be used for automated skin tone extraction at scale. In contrast, ITA can be used to automatically classify skin tone. ITA is a mathematical transformation of skin color in the CIELAB color space, which encodes color as a combination of three values: lightness $L^*$, red-green scale $a^*$ and blue-yellow scale $b^*$. The transformations from RGB to CIELAB color space are well known, and it is recommended \cite{DelBino2013} to use the D65 illumination function in the ITA calculations. Following the transformation of the mean skin tone color from RGB to a CIELAB, the ITA value is determined as
\begin{equation}
ITA = \frac{180}{\pi} \tan^{-1} \left(\frac{L^* - 50}{b^*}\right).
\end{equation}

To perform our test for distributional parity, the continuous ITA scores must be mapped to discrete categories. Although mappings from ITA continuous values to category labels have been proposed by \cite{DelBino2013} as well as \cite{saint2015colour}, both systems use color words (e.g., "intermediate", "golden") that are neither precise nor strictly accurate. Moreover, color terms as applied to skin tone are often imbued with sociocultural meanings that are not related to objective quantification of skin tone \cite{saint2015colour}. To avoid introducing such subjectivity into our category labels, we label the ITA categories as ST1 (i.e., skin tone 1) through ST6 where lower values indicate darker skin tone. Our mappings between ITA values and skin tone categories are presented in in Table \ref{tab:ita-mapping}. 

\begin{table}[htbp]
\caption{Mapping of the ITA values to six categorical skin tone labels \cite{DelBino2013}}
\begin{tabular}{cc}
\toprule ITA, $\deg$ & ITA Skin Tone \\ \midrule
$  ITA < -30 $ & ST1 \\ \hline
$ -30 \leq ITA < 10 $ & ST2 \\ \hline
$ 10 \leq ITA < 28 $ & ST3 \\ \hline
$ 28 \leq ITA < 41 $ & ST4 \\ \hline
$ 41 \leq ITA < 55 $ & ST5 \\ \hline
$ 55 \leq ITA $ & ST6 \\ \bottomrule
\end{tabular}
\label{tab:ita-mapping}
\end{table}

In any image, the ITA values vary across skin pixels. To determine the skin tone category of a model in an image, we use the median ITA value in the images, and map that value to its associated skin tone category.
Figure \ref{fig:segmentation-ita-example} displays examples of skin tone predictions and image segmentation predictions on images from the DeepFashion In-Shop Clothes Retrieval dataset \cite{liuLQWTcvpr16DeepFashion}. The top row of the figure displays the original image and predicted skin tone for fashion models with different ITAs. The values of the ITA are reported above the original image ranging from -33 to 53. The bottom row shows the image segmentation results for the image above where yellow corresponds to pixels labeled as skin. 


\section{Results}

We demonstrate statistical properties of our test using Monte Carlo simulated recommendations and empirical applicability using visual search on the DeepFashion In-Shop Clothes Retrieval \cite{liuLQWTcvpr16DeepFashion} dataset. This dataset is similar to online clothing catalogs as it contains styled images of garments from multiple viewpoints sorted by gender (i.e., Men's vs. Women's) and garment type (e.g., jeans, dresses, etc.). We perform our test for distributional parity globally--across the whole dataset, as well as within a garment type and viewpoint, e.g., front-facing Women's dresses. 

\subsection{Statistical Power with Simulated Data}\label{sec:monte-carlo-results}

\begin{figure*}[htbp]
\centering
\begin{subfigure}[t]{0.28\textwidth}
\includegraphics[width=\linewidth]{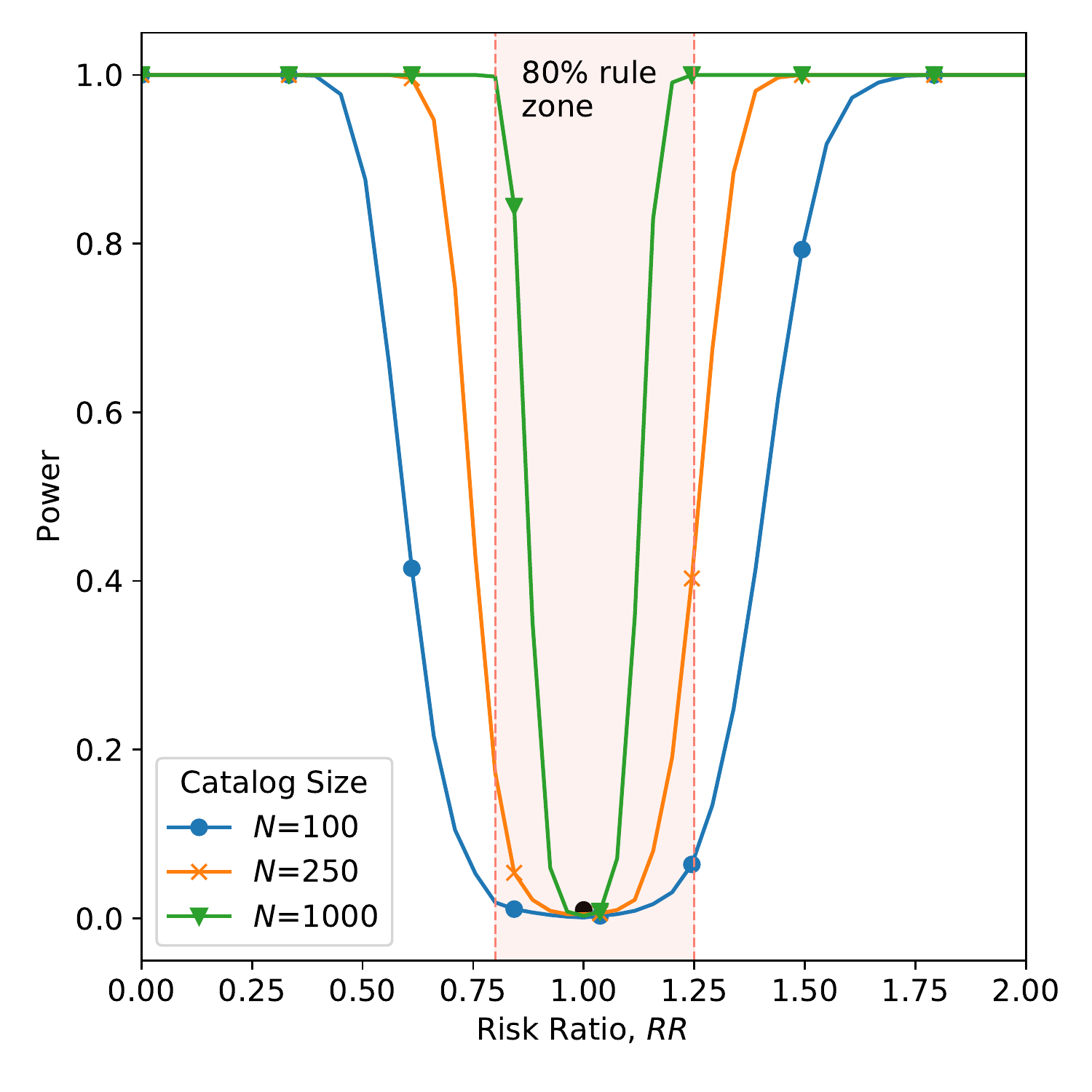}
\caption{Power of the proposed hypothesis test as a function of the risk ratio ($RR$) and catalog size}
\label{fig:global-power}
\end{subfigure}%
\hspace{0.05\textwidth}
\begin{subfigure}[t]{0.28\textwidth}
\includegraphics[width=\linewidth]{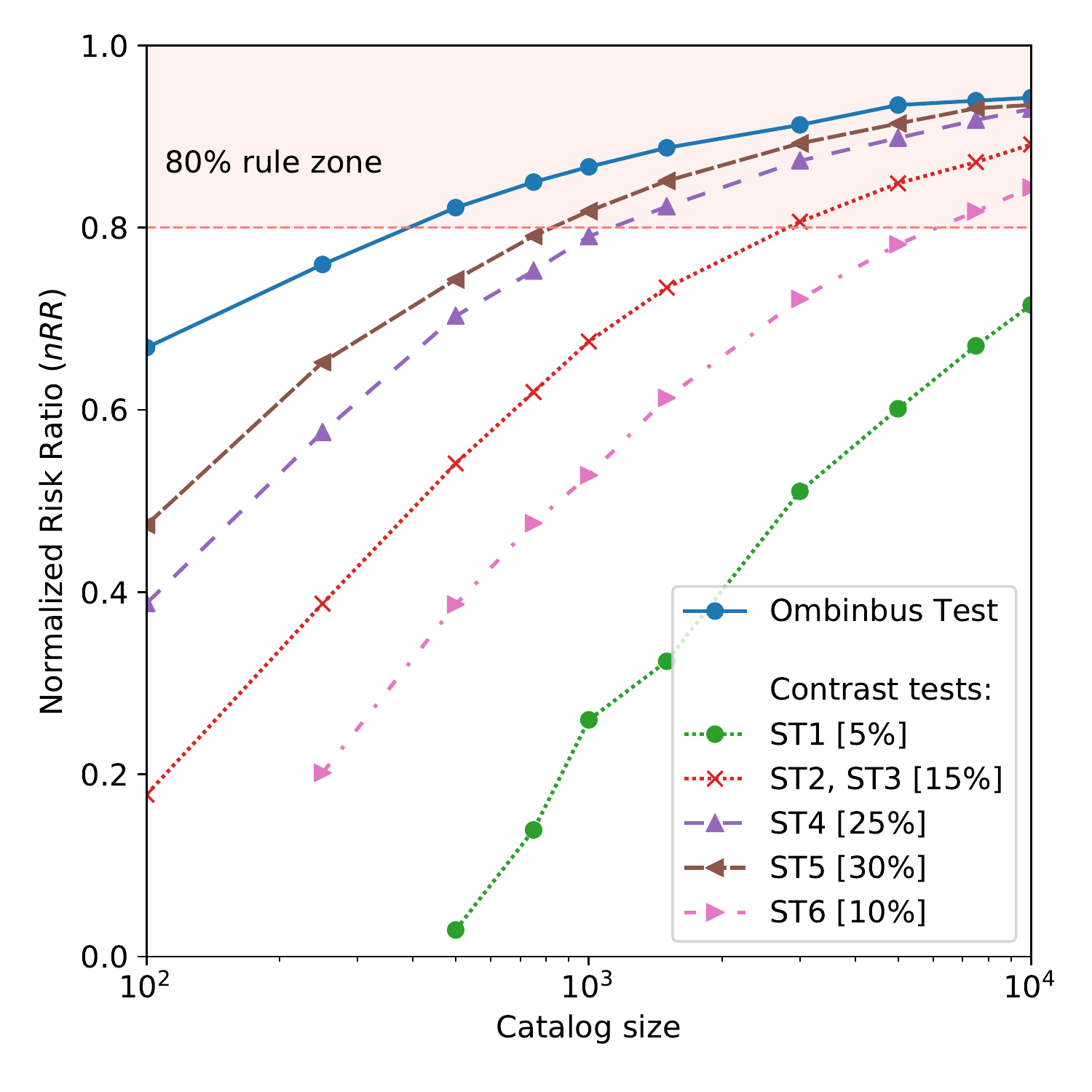}
\caption{Detectable risk ratio as a function of the samples size at 80\% power (RR < 1)}
\label{fig:global-power-sample-size-l}
\end{subfigure}
\hspace{0.05\textwidth}
\begin{subfigure}[t]{0.28\textwidth}
\includegraphics[width=\linewidth]{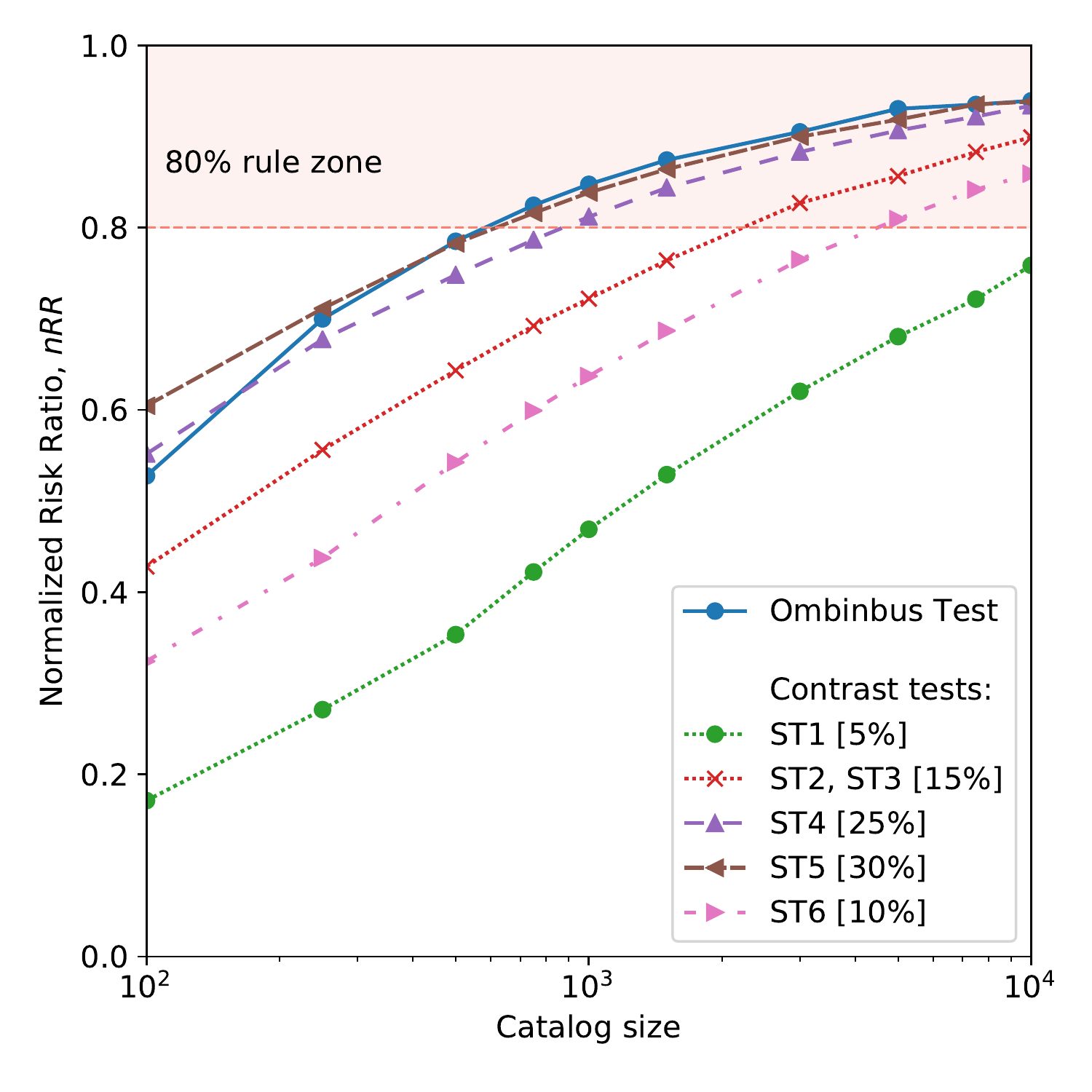}
\caption{Detectable risk ratio as a function of the samples size at 80\% power (RR > 1)}
\label{fig:global-power-sample-size-r}
\end{subfigure}
\caption{(a) The power of the omnibus hypothesis test, $P(\textrm{reject } H_0^O | H_1)$, as a function of the risk ratio for the search catalog size 100, 250, and 1000. (b) The detectable risk ratio for omnibus and contrast tests at 80\% power as a function of the catalog size. Significance level $\alpha < 0.01$ is used when executing tests and calculating the detectable risk ratio.}
\end{figure*}

\begin{table}[htbp]
    \centering
    \caption{Protected variable's distribution of skin tones in Monte Carlo trials}
    \begin{tabular}{cc}
    \toprule
        ITA Skin Tone & Frequency, \%  \\ \midrule
        ST1 & 5 \\
        ST2 & 15 \\
        ST3 & 15 \\
        ST4 & 25 \\
        ST5 & 30 \\
        ST6 & 10 \\\bottomrule
    \end{tabular}
    \label{tab:label-distribution}
\end{table}

The Monte Carlo trials were generated using Algorithm \ref{alg:monte-carlo-sampling} and the protected variable's distribution of skin tones in Table \ref{tab:label-distribution}. When evaluating the hypothesis tests, we utilize the significance level $\alpha<0.01$ unless otherwise specified. The reported trends hold for other values of $\alpha$. For each set of parameters, 1000 Monte Carlo simulations were run to gather sufficient enough data for our analysis.

The power of the omnibus hypothesis test $P(\text{reject } H_0^O |  H_1)$ (see section \ref{sec:test-definitions}) as a function of the risk ratio is displayed in Figure \ref{fig:global-power}. We report the power plots for three catalog sizes, 100, 250, and 1000 observations. When the risk ratio is close to one and the distribution of skin tones in the recommendations are close to the catalog's distribution of skin tones, the test has the least amount of power, achieving the minimum at $RR=1$. The test easily detects large biases, $nRR \ll 1$, for small samples, which is indicated by the value of the test's power being close to 1. The power curve displays an inverted bell shape, becoming narrower for larger catalog sizes. A narrow power curve indicates that the hypothesis test can reliably detect smaller biases, i.e., risk ratios closer to 1. The power curves for the contrast tests show similar characteristics to the omnibus test curves shown in Figure \ref{fig:global-power}.

Figures \ref{fig:global-power-sample-size-l} and \ref{fig:global-power-sample-size-r} display the detectable bias as a function of the catalog size given 80\% power at the significance level $\alpha<0.01$ for $RR<1$ and $RR>1$ respectively. For the omnibus test, both figures indicate that the detectable bias is decreasing, $RR$ getting close to 1, as a function of a catalog size reaching $nRR=0.8$ (80\% rule) at approximately 400-600 samples in the dataset. 


Looking at the contrast tests, it is clear that if a protected value is less represented in the dataset, the number of samples needed to detect an $nRR=0.8$ increases. There is a clear difference between the curves in both figures. When the $RR<1$, the curves drop off more drastically when there is a small catalog size than when the $RR>1$. This is due to the $RR$ having an upper bound that varies with $p(z)$; the upper bound is $\max RR \left(z^q \right) = 1/P_Z(z)$. Therefore, the $nRR$ for $RR>1$ will only approach 0 when $P_Z(z) \to 0$. Also, the detectable bias depends on a skin tone's proportion in the catalog. More specifically, the smaller the proportion a skin tone makes up in a catalog, the larger the bias needed to detect one. This can be seen in Figures \ref{fig:global-power-sample-size-l} and \ref{fig:global-power-sample-size-r} with smaller proportion skin tones having smaller detectable $nRR$'s at the same catalog size.

In the case where the protected variable's distribution of values in the catalog is different from Table \ref{tab:label-distribution}, the methods described in this section can be used to infer the power curves, the detectable risk ratio given the catalog size, and the applicability of the combined omnibus and contrast tests.

\subsection{Visual Search for Fashion Recommendations}
The testing methodology described in Section \ref{sec:approach} is empirically evaluated on a fashion visual search system's recommendations. We perform a search on two sets of images (catalogs) of different sizes, built as subsets of the DeepFashion In-Shop Clothes Retrieval dataset \cite{liuLQWTcvpr16DeepFashion}. The first catalog consists of 51,740 images available in the DeepFashion dataset (i.e., the "full" dataset) which have non-null ITA values, and includes multiple views of the same garments worn on the model. The second catalog is constructed as the frontal views of Women Dresses (i.e., the "dresses" dataset) comprising 1,812 images.

The distribution of the skin tones in the dresses catalog almost matches the distribution of the skin tones in the full catalog (see Table \ref{tab:ita-distribution-deepfashion}). In both cases, the ST6, ST5 and ST4 skin tones compose the majority of the images accounting for 83\% and 88\% of the full and the dresses subset, respectively. The ST1, ST2, and ST3 skin tones are underrepresented in both subsets, having merely 234 and 2 images associated with the ST1 skin tone label in the full and dresses catalogs, respectively.

\begin{table}[htbp]
 \caption{Distribution of the ITA skin tone labels extracted from the DeepFashion In-Shop Clothes Retrieval dataset \cite{liuLQWTcvpr16DeepFashion}. The distributions are reported in the full catalog and the subset of restricted to the Women's Dresses}
\begin{tabular}{ccc}
\toprule 
ITA Skin Tone & \multicolumn{2}{c}{Frequency, \%} \\ \cline{2-3}
& Full Catalog & Dresses Catalog\\ \midrule
ST1 & 0.4 & 0.1\\ \hline
ST2 & 3.7 & 1.7 \\ \hline
ST3 & 11 & 10 \\ \hline
ST4 & 25 & 26\\ \hline
ST5 & 40 & 46\\ \hline
ST6 & 18 & 16\\ \hline
\end{tabular}
\label{tab:ita-distribution-deepfashion}
\end{table}

A bias is detected in the fashion recommendations generated using the approach described in \ref{sec:visual-search} with $K$=6 nearest neighbors. The omnibus test detects a bias on both the full and dresses catalogs. Also, the contrast test detects a bias on the full catalog and only a limited set of the protected values for the dresses catalog (see Table \ref{tab:contrast-deepfashion}). Thus, the contrast test rejects $H_0^C(z^q)$ only for $z^q=$ ST3 \ at $\alpha = 10^{-3}$ on the women dresses search, a smaller catalog. A bias is detected for all of the values of the query skin tone on a large catalog search, highlighting that a large number of samples is required to detect smaller biases on smaller catalogs. For example, the $RR(z^q=\text{ST2})$ is estimated at 2.01, however, the bias is not detected given that there are only 30 query items, which is not sufficient to detect a bias in their recommendations.

The strongest bias is estimated for the fashion images associated with the models of the ST1 (i.e., darkest) skin tone, registering at $RR=8.41$. Notably, the bias on the dresses search for the same values of the protected variable is detected by a $\chi^2$ test; however, the confidence interval of the risk ratio contains $RR=1$. Having only 2 query images with a model that have a ST1 skin tone, the disagreement between risk ratio confidence interval and $\chi^2$ test results indicates lack of the statistical power to detect bias. The risk ratios estimated from the search performed on the full and dresses catalogs are closely aligned. Investigating whether the bias is an intrinsic property of the search algorithm used is outside of the scope of this paper and requires further exploration.

\begin{table}[htbp]
\centering
\caption{Results of the omnibus test for distribution parity in skin tone from recommendations made with $K=6$ nearest neighbors}
\begin{tabular}{ccc}
\toprule
Catalog     & size      & $\chi^2(30)$ \\ \midrule
Full        &   51740   & $16627^{***}$ \\
Dresses     &   1812    & $304^{***}$ \\ \bottomrule
\multicolumn{3}{l}{\footnotesize{\textsuperscript{*} $p < 10^{-2}$;} \footnotesize{\textsuperscript{**} $p < 10^{-3}$;} \footnotesize{\textsuperscript{***} $p < 10^{-4}$;}} \\
\end{tabular}
\label{tab:omnibus-deepfashion}
\end{table}

\begin{table}[htbp]
\small
\centering
\caption{Results of contrast tests for distribution parity in skin tones from recommendations made with $K=6$ nearest neighbors}
\begin{tabular}{ccccc}
\toprule
\shortstack{Protected variable \\in the query, $z^q$}
& Catalog & Size & $\chi^2(1)$ & $RR[95\% CI]$\\	\midrule
\multirow{2}{*}{ST1} & Full & 234 & $378^{***}$ & 8.41 [6.54, 10.8] \\
& Dresses & 2 & $11.9^{**}$ & 0 [0, $\infty$] \\ \hline

\multirow{2}{*}{ST2} & Full & 1963 & $1054^{***}$ & 2.79  [2.53, 2.87] \\
& Dresses & 30	& $2.74$ & 2.01 [0.97, 4.19] \\ \hline

\multirow{2}{*}{ST3} & Full & 5904 & $786^{***}$ & 1.54 [1.49, 1.58] \\
& Dresses & 182 & $20.5^{***}$ & 1.51 [1.27, 1.81] \\ \hline

\multirow{2}{*}{ST4} & Full & 13121 & $274^{***}$ &  1.15 [1.13, 1.17]	\\
& Dresses & 469 & 3.76 & 1.10 [1.00, 1.20] \\ \hline

\multirow{2}{*}{ST5} & Full & 21123 & $301^{***}$ &  1.10 [1.09, 1.12]	\\
& Dresses & 821 & $7.41^{*}$ & 1.08 [1.02, 1.14] \\ \hline

\multirow{2}{*}{ST6} & Full & 9395 & $867^{***}$ &  1.38 [1.35, 1.41] \\
& Dresses & 298 & $9.27^{*}$ & 1.22 [1.07,1.39] \\  \bottomrule
\multicolumn{3}{l}{\footnotesize{\textsuperscript{*} $p < 10^{-2}$;} \footnotesize{\textsuperscript{**} $p < 10^{-3}$;} \footnotesize{\textsuperscript{***} $p < 10^{-4}$;}} \\
\end{tabular}
\label{tab:contrast-deepfashion}
\end{table}

\section{Limitations}

\begin{figure}
    \centering
    \includegraphics[width=\linewidth]{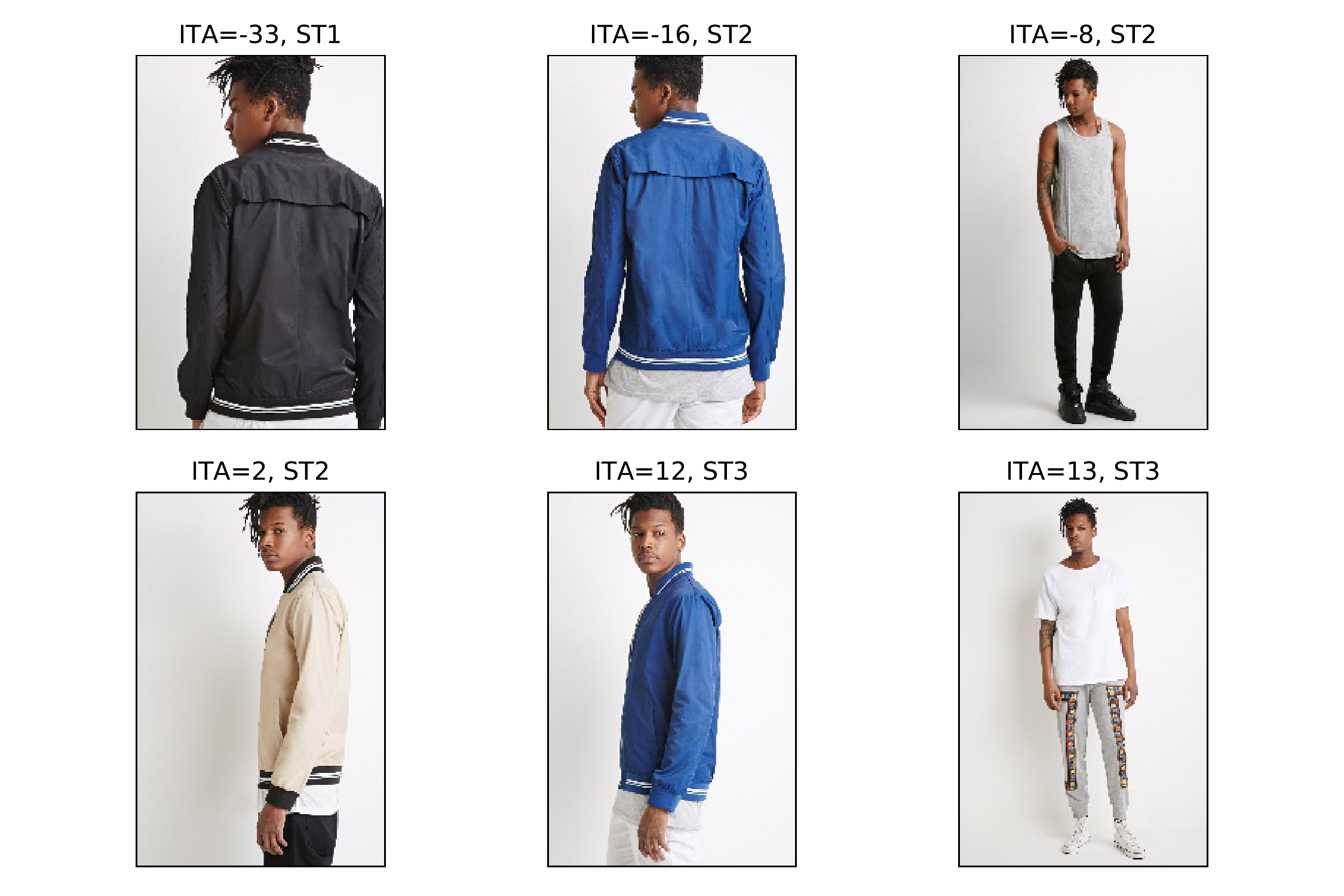}
    \caption{Variation in ITA values vary for the same model. Estimated standard deviation of $\sigma_{ITA} = 6.2$.}
    \label{fig:ita-variation}
\end{figure}

Several technical and practical limitations of our approach are worth noting. First, our categorization of skin tone using ITA is a measure of skin tone within the context of a particular image rather than a fixed, objective measure of skin tone. The visual appearance of colors in images depends on the lighting, shadows, make-up, etc., which can vary the values of the ITA for the same fashion model. Figure \ref{fig:ita-variation} shows the ITA values generated for six images of the same model ranging from -33 to 13, which corresponds to ST1, ST2, and ST3 skin tones. The standard deviation of the ITA values for a single model is estimated from multiple views of the same product is $\sigma_{ITA} = 6.2$, assuming that only one model is present in the set of product images. This estimate of $\sigma_{ITA}$ is conservative because the sample of images containing the same model is not readily available in the DeepFashion dataset. Thus, the ITA values and generated labels should be interpreted as color features that capture visually-apparent skin tone rather than the true skin tone of a model.

Second, in our power analysis, we leverage the 80\% rule, which represents a legal standard for whether or not an algorithm's output is biased \cite{zafar2015fairness}. Having a single, static threshold for bias is practically useful. With a static threshold, we can build automated tests to determine if recommendations merit manual review. However, using the 80\% rule to set a threshold is problematic for functional as well as statistical reasons. Within the recommendation domain, there is no evidence to suggest that the level at which humans perceive skin tone and/or racial bias in algorithms' recommendations is consistent with the thresholds that correspond to the 80\% rule. Moreover, the perceptual threshold for perceiving skin tone or racial bias in recommendations may depend both on the domain of application (e.g., fashion vs. cosmetics) as well as characteristics of users (e.g., race, gender). For example, being a racial minority and having previously experienced subtle forms of racial bias is associated with an increased likelihood of perceiving  racially-charged internet memes as offensive \cite{williams2016racial}. Our power analysis demonstrates the statistical complexities of using a fixed point estimate as a threshold---whether a bias can be statistically detected depends on properties of the sample. A more valid method for setting a functionally relevant threshold would involve conducting user testing in the specific domain of application within multiple user groups to determine the level at which users detect bias.

Also, the test we propose here involves performing multiple statistical tests; therefore, we are more likely to commit a Type I error with each test we conduct. Yet, we do not propose a specific multiple comparisons correction (MCP) here for several reasons. The choice of MCP will depend on whether the tests are planned or post-hoc, simple or complex comparisons, and whether there are many or only a few comparisons. While a Bonferroni correction may be reasonable for an analysis with sufficiently few follow-up comparisons, a Bonferroni correction would be overly conservative if sufficiently many comparisons are performed \cite{field2012discovering}. The choice of MCP also depends on whether the practitioner is more concerned with controlling the Type I or Type II error rate. If our test is used to identify potentially biased recommendation results for manual review, having numerous Type I errors would create unnecessary manual labor; however, by stringently controlling the Type I error rate, we increase our chances of committing a Type II error. That is, we are more likely to fail to flag some biased results. Practitioners who are more interested in reducing the chance of serving biased results to users may do well to consider a less conservative MCP such as those based on false discovery rate.

Furthermore, our proposed method is focused on identifying rather than correcting deviations from distribution parity. Identifying bias is not equivalent to providing unbiased results. That is, our method does not specify what to  do in the event that a bias is detected. Lipton et al. \cite{lipton2018does} demonstrate that well-intentioned attempts to render algorithmic outcomes fair can sometimes result in harm to particular individuals within a disadvantaged group. Therefore, proposed methodologies for correcting recommendations with a bias should carefully consider any potential negative consequences of the correction. Although proposing specific guidelines for correcting bias is outside the scope of the current paper, we have observed that we can significantly reduce the extent to which visual search recommendations are biased by skin tone. One approach is to use segmentation models trained on images featuring more diverse fashion models and use these segmentations to remove skin pixels from images before performing similar item retrieval. Another is to use embeddings generated by fashion specific classifiers trained again on diverse data-sets of fashion models. Future work should examine these and other methodologies for providing search results that are independent of protected variables.   

Lastly, our empirical analysis is limited in its usefulness due to issues of representation in the data set used. First, although our effect sizes for ST1 (i.e., the darkest skin tone) were the largest observed, our ability to detect these effects is severely constrained by the low number of fashion models with an ST1 skin tone. The low number of fashion models with darker skin tones is not limited to the DeepFashion dataset. In an analysis of cover models in \textit{Vogue Magazine} over the last few years, Handa \cite{handa_2019} demonstrates that although the magazine has featured more models of color in recent years, still very few have a darker skin tone. As a result, our test is least likely to detect bias in the most marginalized group of fashion models. Recently, the fashion industry has responded to calls for increased inclusivity by featuring fashion models with more racial, cultural, age, and body shape diversity \cite{Day2018}. However, without greater diversity in human models in fashion images, attempts to detect bias along any single protected variable, let alone intersecting protected variables, will be limited. 

\section{Discussion}


The goal of the current work was to develop a test for detecting fairness in IR systems. Here we describe our test for distribution parity, which determines whether the presence of a protected variable value in a query affects the likelihood that the resulting recommendations will also share that value. Although the distribution parity test could be used for a range of protected variables with categorical values, we chose to evaluate our test for distribution parity using skin tone bias in an image-based fashion IR system as an example use case. To demonstrate the utility of the test within this context, we performed an evaluation on a publicly available dataset using the 80\% rule to set our bias threshold. In the DeepFashion full dataset, which has a sufficient number of samples per skin tone category, our test reveals a statistically significant bias in recommendations. Also, the DeepFashion dresses dataset had a statistically significant bias, but it was not possible to conclude that there was bias in certain skin tones because of an insufficient number of images. Through these results, we have shown that our method for detecting bias can be a powerful tool for ensuring that users are given high quality, unbiased results, but also that this test cannot find statistically significant bias when some skin tone categories have limited representation in a dataset. 

As recommender and IR systems become more prevalent, it will be increasingly important to develop methods for determining if a system outputs are biased. Bias in algorithms informing high-stakes decision making is straightforwardly damaging to some user segments. For example, \textit{ProPublica} \cite{angwin2016} showed that one widely employed recidivism prediction system falsely predicted higher rates of recidivism among black than white defendants. Yet, bias in an algorithms' predictions in seemingly benign contexts such as fashion still merit investigation. If factors independent of the recommendation domain bias algorithms’ results--i.e., if skin tone influences fashion recommendations---the resulting recommendations will necessarily provide a poor experience for users. Beyond contributing to bad user experience, results with bias could also have deleterious effects on marginalized individuals, who already regularly experience bias in their day-to-day lives. For example, experiencing racially biased results from a recommender system is conceptually similar to experiencing other racial microaggressions--defined as subtle, daily experiences that intentionally or unintentionally insult, degrade, or invalidate racial minorities  \cite{Wong2014}. In racial minorities, individuals who report having experienced microaggressions also report poorer physical, mental, and occupational outcomes \cite{Wong2014}. Therefore, providing recommendations that manifest skin tone and/or racial biases could contribute to the constellation of negative experiences marginalized people frequently experience. Similarly, if IR systems are not fair with regard to other dimensions of users identities such as gender (see \cite{barthelemy2016gender, basford2014you}) or intersecting identities (see \cite{nadal2015qualitative}), many users may be especially impacted by IR systems in negative ways.

The method we propose here offers an avenue for understanding bias within the recommendation domain. Although there are some caveats for its application, our approach can help ensure that all users are provided with a high quality recommendation experience.


\newpage
\bibliographystyle{ACM-Reference-Format}
\bibliography{library.bib}

\appendix

\end{document}